\documentclass[10pt,a4paper,twocolumn,twoside]{article}
\usepackage{amsmath}
\usepackage{amssymb}
\usepackage[utf8]{inputenc}
\usepackage{forest}
\usepackage{hyperref}
\usepackage{makecell}
\usepackage{float}
\usepackage[section]{placeins}
\hypersetup{
    colorlinks=true,
    linkcolor=blue,       
    citecolor=blue,       
    filecolor=magenta,    
    urlcolor=cyan,
    pdftitle={An Attention-Augmented VAE-BiLSTM Framework for Anomaly Detection in 12-Lead ECG Signals},
    pdfauthor={Marc Garreta Basora},
    pdfsubject={Anomaly Detection in ECGs},
    pdfkeywords={Anomaly Detection, ECG, Autoencoder, Variational Autoencoder, Attention Mechanism, Unsupervised Learning, Visualization Dashboard}}
\usepackage{graphicx}
\usepackage{times}
\usepackage{titlesec}
\usepackage{multirow}
\usepackage{lettrine}
\usepackage{booktabs}
\usepackage[top=2cm, bottom=1.5cm, left=2cm, right=2cm]{geometry}
\usepackage[figurename=Fig.,tablename=TABLE]{caption}
\usepackage[affil-it]{authblk}

\title{\LARGE\sffamily An Attention-Augmented VAE-BiLSTM Framework for \\Anomaly Detection in 12-Lead ECG Signals}

\author[1*]{Marc Garreta Basora}
\author[2,1]{Mehmet Oguz Mulayim}
\affil[1]{Universitat Autònoma de Barcelona (UAB), Cerdanyola del Vallès, Spain}
\affil[2]{Artificial Intelligence Research Institute (IIIA-CSIC),
Cerdanyola del Vallès, Spain}

\date{}

\newcommand\blfootnote[1]{%
  \begingroup
  \renewcommand\thefootnote{}\footnote{#1}%
  \addtocounter{footnote}{-1}%
  \endgroup
}

\titleformat{\section}
{\large\sffamily\scshape\bfseries}
{\textbf{\thesection}}{1em}{}

\makeatletter
\renewenvironment{abstract}{%
  \begin{center}%
    \sffamily\normalfont\Large\bfseries Abstract
  \end{center}%
  \vspace{-0.5em} 
}{}
\makeatother

\begin{document}
\twocolumn[
%\begin{@twocolumnfalse}

\maketitle  
\begin{center}
\parbox{0.915\textwidth}
{\small\sffamily
\begin{abstract}
Anomaly detection in 12-lead electrocardiograms (ECGs) is critical for identifying deviations associated with cardiovascular disease. This work presents a comparative analysis of three autoencoder-based architectures: convolutional autoencoder (CAE), variational autoencoder with bidirectional long short-term memory (VAE-BiLSTM), and VAE-BiLSTM with multi-head attention (VAE-BiLSTM-MHA), for unsupervised anomaly detection in ECGs. To the best of our knowledge, this study reports the first application of a VAE-BiLSTM-MHA architecture to ECG anomaly detection. All models are trained on normal ECG samples to reconstruct non-anomalous cardiac morphology and detect deviations indicative of disease. Using a unified preprocessing and evaluation pipeline on the public China Physiological Signal Challenge (CPSC) dataset, the attention-augmented VAE achieves the best performance, with an AUPRC of 0.81 and a recall of 0.85 on the held-out test set, outperforming the other architectures. To support clinical triage, this model is further integrated into an interactive dashboard that visualizes anomaly localization. In addition, a performance comparison with baseline models from the literature is provided.
\end{abstract}
\bigskip

}
\bigskip

{\vrule depth 0pt height 0.5pt width 4cm\hspace{7.5pt}%
\raisebox{-3.5pt}{\fontfamily{pzd}\fontencoding{U}\fontseries{m}\fontshape{n}\fontsize{10}{10}\selectfont\char70}%
\hspace{7.5pt}\vrule depth 0pt height 0.5pt width 4cm\relax}
\end{center}
%]
\bigskip
]

%\end{@twocolumnfalse}

\blfootnote{$^*$ Corresponding author: \texttt{Marc.GarretaB@autonoma.cat}}

\section{Introduction}
\label{sec:introduction}
Anomaly detection (AD) refers to the process of identifying patterns that deviate from an expected or normal behavior in the data \cite{01_ad_definition}. The importance of this data identification lies in the fact that these variations, known as anomalies, may lead to critical actionable information \cite{02_ad_definition}. For instance, AD plays a crucial role in the healthcare domain, as they can, when properly processed, indicate potential diseases or critical health events in patients. More specifically, cardiovascular diseases (CVDs) are the leading cause of death globally \cite{ref1}, accounting for over one third of all deaths every year \cite{ref2}. Early detection of cardiac issues is therefore essential, as it improves patients' quality of life by providing early warning of upcoming health events, reduces the economic burden on healthcare systems, and even saves lives \cite{ref7}. Among various diagnostic tools, the electrocardiogram (ECG) is one of the most common, non-invasive methods that is used as a diagnostic tool. Because it records key information such as heart rhythm, heart rate and cardiac axis information \cite{ref3}, the ECG is an important data for early recognition of various cardiac conditions such as coronary artery disease (CAD), heart failure (HF), arrhythmia (ARR), and other heart diseases. 

This time-series signal captures the electrical activity of the heart, reflecting how electrical impulses propagate through cardiac tissues and can be detected via electrodes placed on the skin \cite{ref5}. In this work, we will focus on the 12-lead ECG, which provides different information from various parts of a patient's body.

Regardless of its diagnostic importance, interpreting multi-lead ECG samples is time-consuming, making even trained physicians misclassify subtle variations that can lead to health diseases \cite{ref7,ref6}. This limitation motivates the development of an automated approach that can assist medical professionals by reducing the time of detection, as well as possibly enhancing the accuracy of identifying cardiac anomalous patterns from ECG data.

\begin{figure}[!ht]
  \centering
  \includegraphics[width=0.3\textwidth]{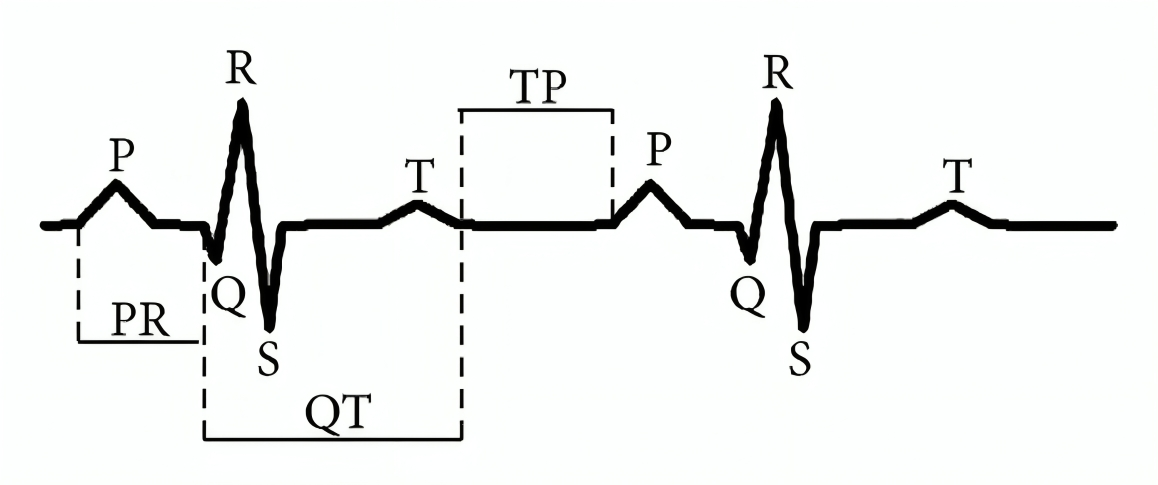}
  \caption{ECG morphology of two normal beats. Reproduced from Zhang et al. \cite{zhang2015robust} with permission.}
  \label{fig:normal_beat}
\end{figure}

To understand these anomalies, it is important to understand the fundamental ECG waveform components. As shown in Figure~\ref{fig:normal_beat}, a normal ECG consists of different intervals and waves that represent the electrical activity of the heart. First, there is the P-wave, continued by the QRS complex, which reflects the contraction of the ventricles when the heart pumps. After that, the T-wave indicates the heart recovery. There are also intervals such as PR, QT or TP, which provide information about the electrical signals flow.\\

In recent research, machine learning (ML) has been applied to anomaly detection across various domains. In the healthcare context, ML models show strong capabilities in processing large volumes of physiological multivariate time-series data, enabling those architectures to learn complex patterns that may be difficult to detect and time-consuming through manual professional inspection. In particular, unsupervised and self-supervised approaches are gaining prominence due to the scarcity of labeled data in this domain.

Among the proposed ML architectures, variational autoencoder (VAE)-based architectures are one of those that have proved effective in distinguishing between normal and abnormal 12-lead ECG recordings \cite{liu2021vqvae, jang2021cvae_ecg}. Additionally, adding attention mechanisms into these architectures has also demonstrated significant benefits in other anomaly detection application domains \cite{correia2023mavae}.

Particularly, this work focuses on autoencoder-based architectures to detect anomalies in multivariate ECG time-series data. Specifically, three models are implemented and compared for the task: the Convolutional Autoencoder (CAE), the Variational Autoencoder with Bidirectional Long Short-Term Memory (VAE-BiLSTM), and the VAE-BiLSTM with Multi-Head Attention (VAE-BiLSTM-MHA).

\subsubsection*{Contributions}

This paper, to the best of our knowledge, proposes the first application of an architecture that integrates Multi-Head Attention into a VAE-BiLSTM framework for 12-lead ECG anomaly detection. The contributions of this work are as follows:
\begin{itemize}
    \item We present a novel approach to 12-lead ECG anomaly detection by using a VAE-BiLSTM-MHA framework, which integrates multi-head attention into a variational autoencoder model. 
    \item We conduct a comparative analysis with custom implementations of two other autoencoder-based architectures: CAE and VAE-BiLSTM.
    \item We develop an interactive dashboard that provides visually interpretable results based on model decisions. 
    \item We provide all code publicly for further research. 
\end{itemize}

\subsubsection*{Document Structure}
The remainder of this paper is structured as follows: Section~\ref{sec:relatedwork} briefly surveys classical and deep learning techniques for ECG anomaly detection. Section~\ref{sec:methodology} describes the methodology followed, including the problem formulation in Section~\ref{sec:problem}, the preprocessing pipeline in Section~\ref{sec:data_preprocessing}, the model architectures in Section~\ref{sec:model} and the anomaly detection strategy in Section~\ref{sec:anomaly_detection}. Section~\ref{sec:experiments} presents the experimental setup and results, covering the used datasets in Section~\ref{sec:datasets}, the applied evaluation metrics in Section~\ref{sec:evaluation_metrics}, the results in Section~\ref{sec:results}, and the interactive dashboard in Section~\ref{sec:dashboard}. Finally, Section~\ref{sec:conclusions} summarizes the conclusions and outlines future work.

\section{State of the Art}
\label{sec:relatedwork}

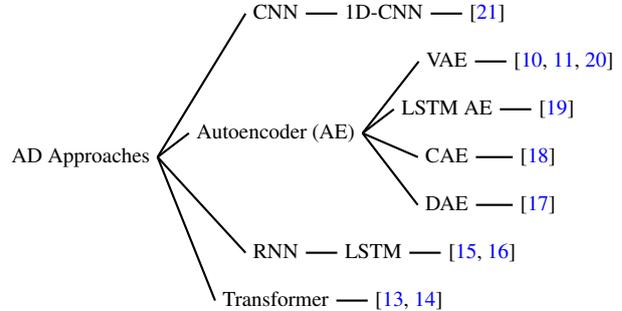
\begin{figure}[htbp]
\centering
\scalebox{1.0}{ % adjust scaling factor as needed
\begin{forest}
for tree={
    grow=east,
    parent anchor=east,
    child anchor=west,
    edge={draw, thick},
    align=left,
    font=\footnotesize, % smaller font
    l=1.2cm,            % shorter branch length
    inner sep=0.5mm,         
    outer sep=0.5mm,
    s sep=1mm           % reduced sibling separation
}
[AD Approaches
    [Transformer
        [\cite{soa_transformer1,soa_transformer2}]
    ]
    [RNN
        [LSTM
            [\cite{chauhan2015anomaly, thill2019anomaly}]
        ]
    ]
    [Autoencoder (AE)
        [DAE
            [\cite{rw_dae}]
        ]
        [CAE
            [\cite{rw_convae}]
        ]
        [LSTM AE
            [\cite{rw_ae_peak_detection}]
        ]
        [VAE
            [\cite{liu2021vqvae, jang2021cvae_ecg, rw_vae_ae}]
        ]
    ]
    [CNN
        [1D-CNN
            [\cite{gu2023lightweight}]
        ]
    ]
]
\end{forest}
}
\caption{Approaches for Anomaly Detection (AD) in ECGs}
\label{fig:ad_tree}
\end{figure}

The field of anomaly detection in healthcare has evolved over time. Early methods relied on rule-based systems \cite{rw_rule_based} and small-scale medical corpora, which were limited when handling the complexity of multivariate time-series data such as ECG or electroencephalogram (EEG) signals. With the increase in data from electronic health records (EHRs) \cite{theodorou2023halo} and advances in deep learning techniques, these latest approaches have overcome the previous ones.

This literature review focuses mostly on unsupervised and self-supervised learning as these methods are considered to tackle the lack of labeled anomalous data in ECG databases \cite{jiang2024anomaly, qin2023novel}. 

Figure \ref{fig:ad_tree} illustrates the anomaly detection approaches that used ECGs as data. Gu et al.\ \cite{gu2023lightweight} design a CNN with depthwise convolutions and 8-bit quantization, specifically adapted for Field-Programmable Gate Array (FPGA) deployment in wearable devices. Their implementation processes 4-second ECG windows in real time, achieving 97.69 \% accuracy on the MIT-BIH \cite{mitdb} arrhythmia dataset and demonstrating that high‐fidelity anomaly detection pipelines can be applied under power and area constraints.

Recurrent architectures have also been widely explored for ECG anomaly detection. Chauhan and Vig \cite{chauhan2015anomaly} apply a deep Long Short-Term Memory (LSTM) to raw ECG time signals, learning temporal representations that identify irregular beats with high detection accuracy. Additionally, Thill et al.\ \cite{thill2019anomaly} propose a stacked LSTM approach that models the multivariate prediction‐error distribution across the ECG leads via a Gaussian model, achieving high recall with low false‐alarm rates on the MIT-BIH arrhythmia dataset.

Autoencoder (AE)–based approaches have shown promising results regarding the reconstruction of normal ECG morphology and detection of anomalies through reconstruction errors. Particularly, variational autoencoders (VAEs) have demonstrated to be effective for ECG anomaly detection \cite{liu2021vqvae, jang2021cvae_ecg, rw_vae_ae}. Liu et al. \cite{liu2021vqvae} used  a vector quantized VAE (VQ-VAE) to perform synthetic data augmentation in order to classify between several cardiac anomalies. Jang et al. \cite{jang2021cvae_ecg} applied a convolutional autoencoder trained with unlabeled data to extract ECG features. Atamny et al. \cite{rw_vae_ae} employed a variational autoencoder, which outperformed the other unsupervised models they evaluated. 
However, the integration of multi-head attention mechanisms within VAE-based models remains unexplored in the context of ECG data. This apparent gap is one of the motivations of this study to explore this autoencoder variant within other architectures, providing a comparative analysis of its performance against similar approaches.

Among other autoencoder approaches, several studies are relevant. Lomoio et al.\ \cite{rw_convae} propose a 1D convolutional autoencoder trained on synthetic ECG segments to learn ``normal'' patterns, reporting AUROC (Area Under the ROC Curve) of 97.82 \% on simulated data and a AUROC of 0.80\% on the CPSC-2018 12-lead ECG test set \cite{cpsc}. They also provide reconstruction‐error heatmaps over input data for explainability, validated against cardiologist annotations. Choi et al.\ \cite{rw_ae_peak_detection} introduce a segment‐wise LSTM autoencoder that processes PreQ, QRS, and PostS intervals separately---corresponding to atrial conduction, ventricular depolarization, and ventricular repolarization phases of the heart’s cycle, respectively---achieving AUROCs up to 0.96 per segment and an overall Atrial Fibrillation (AF) detection AUROC of 0.98 when the three anomaly scores are fused via an XGBoost classifier, and report a AUROC of 0.74\% on the CPSC-2018 12-lead ECG test set. Hribar and Torkar \cite{rw_dae} develop a denoising autoencoder for 12-lead ECG that removes the need for band-pass filters---which limit the frequency of the ECG signal--- and notch filters---which remove narrow-band interference---, attaining 0.81 accuracy and 0.74 recall on the PhysioNet/CinC 2021 challenge dataset \cite{reyna2021will}, with saliency overlays pinpointing the temporal origins of anomalies. In a comprehensive comparative study, Atamny et al.\ \cite{rw_vae_ae} benchmark standard AEs, VAEs, diffusion models, normalizing flows, and Gaussian mixture models on the CPSC-2018 12-lead ECG challenge, finding the VAE as the best performant with AUROC = 0.83 while even the simplest AE achieves AUROC = 0.76.

Regarding the utilization of VAE-based models in other domains , Fu et al.\ \cite{OmniAnomaly} proposed OmniAnomaly, a VAE-based architecture enhanced with adversarial training and probability reconstruction generation in industrial sensor streams. Correia et al.\ \cite{correia2023mavae} introduced MA-VAE, which integrates multi-head attention into the VAE framework, showing promising results results in complex temporal datasets in Automotive Endurance Powertrain Testing.

Transformer-based architectures are also used in the task of anomaly detection and, this variety of models excel at modeling multivariate time-series data-ECG data-by capturing both temporal and spatial dependencies between multiple leads and its duration-timesteps. For instance, Hu et al. \cite{soa_transformer1} presents a hybrid CNN-Transformer network that achieves state-of-the-art arrhythmia classification on single-lead data by combining local feature extraction with global self-attention. Similarly, Kim et al.  \cite{soa_transformer2} introduces S-transform–augmented CNN–Transformer that preprocesses the input into a time–frequency representation before attention pooling, producing an improvement when detecting subtle waveform anomalies. In parallel, Tuli et al. \cite{TranAD} proposes TranAD, a Transformer-based model that captures long-term dependencies for robust anomaly detection.\\

\noindent Motivated by the high performance of autoencoder-based ECG anomaly detection systems, as explained above, the absence of any study implementing multi-head attention in VAE architectures for ECG data emphasizes the novelty of this work and encourages its comparative evaluation against similar models. 

\section{Methodology}
\label{sec:methodology}

This section describes the main steps taken to design and implement the anomaly detection systems for 12-lead ECG signals, including the formulation of the problem, data preprocessing, model development, and the anomaly detection process.

\subsection{Problem Formulation}
\label{sec:problem}

A multivariate time series is a sequence of data points along \( m \) dimensions. In the context of ECG signals, each dimension corresponds to one of the 12 leads, resulting in \(m=12\). Therefore, a 12-lead ECG dataset can be represented as a multivariate time series: 

\begin{equation}
    \mathcal{T} = \{ \mathbf{x}_1, \mathbf{x}_2, \ldots, \mathbf{x}_T \}, \quad \mathbf{x}_t \in \mathbb{R}^{m} \quad \text{\cite{TranAD}}
\end{equation}

\noindent where each observation \( \mathbf{x}_t \in \mathbb{R}^{12} \) is composed by all 12 leads at time  
 \( t \).

\noindent In this unsupervised learning setting, the autoencoder-based models are trained exclusively on normal samples $\mathcal{T}$ to basically learn a compact representation of the healthy signal manifold and to overcome the imbalance between normal and anomalous ECG samples in clinical corpora \cite{jiang2024anomaly, qin2023novel}. During inference or testing, given an unseen sample \( \hat{\mathbf{x}}_t \), the task is to compute how much a 12-lead ECG test sample deviates from the learned representation to decide whether that sample diverges from the normal-signal manifold or not. That is, if the unseen sample lies too far from the normal representation, it could be considered an anomaly.

\noindent To measure this difference, an anomaly score \( S_t \) is defined, which is compared against a threshold \( \tau \) to assign a binary anomaly label \( y_t \):

\begin{equation}
y_t =
\begin{cases}
1, & \text{if } S_t > \tau, \\
0, & \text{otherwise},
\end{cases}
\quad\cite{chauhan2015anomaly}
\label{eq:threshold}
\end{equation}

The value of \( S_t\) is computed per each window along each of the 12th leads, so at the end, you end up with several scores per lead that are averaged to obtain a single score per sample (See \ref{sec:window_segmentation} for more details regarding the Windowing process).

The value of \(\tau\) is not fixed but estimated on the validation split that contains only normal ECG recordings. Four strategies for deciding the adequate threshold between a normal representation and an anomalous sample are explored:

\begin{enumerate}
  \item \textbf{95th percentile.}  It is an unsupervised rule that its threshold is fixed at the 95th percentile of the validation scores from normal samples.

  \item \textbf{F\(_1\)-optimisation.} This approach keeps the value \(\tau\) that maximises the F\(_1\) score during validation. 

  \item \textbf{Youden’s \(J\) statistic \textnormal{\cite{Youden}}.} This technique optimizes the Receiver Operating Characteristic (ROC) curve during validation, selecting the threshold that maximizes \(J=\mathrm{TPR (True PositiveRate)}-\mathrm{FPR(FalsePositiveRate)}\). 

  \item \textbf{Peaks-Over-Threshold (POT) \textnormal{\cite{OmniAnomaly,POT}}.}  It uses one of the above techniques to choose a baseline threshold \(u\). Then, a Generalised Pareto Distribution \cite{pareto} is fit to the extreme tail of the validation scores to set \(\tau\). 

\end{enumerate}

In this study, the above procedure is repeated for every architecture
(i.e., CAE, VAE-BiLSTM, VAE-BiLSTM-MHA), obtaining a model-specific threshold that is used during testing with the goal of minimizing the ``false alarm'' \cite{false_alarm} principle, a concept used in the medical domain that occurs when a system erroneously classifies a normal sample as anomalous. 

\subsection{Data Preprocessing}
\label{sec:data_preprocessing}

\subsubsection{Input Data}
\label{sec:input_data}

\begin{figure}[!ht]
  \centering
  \includegraphics[width=0.4\textwidth]{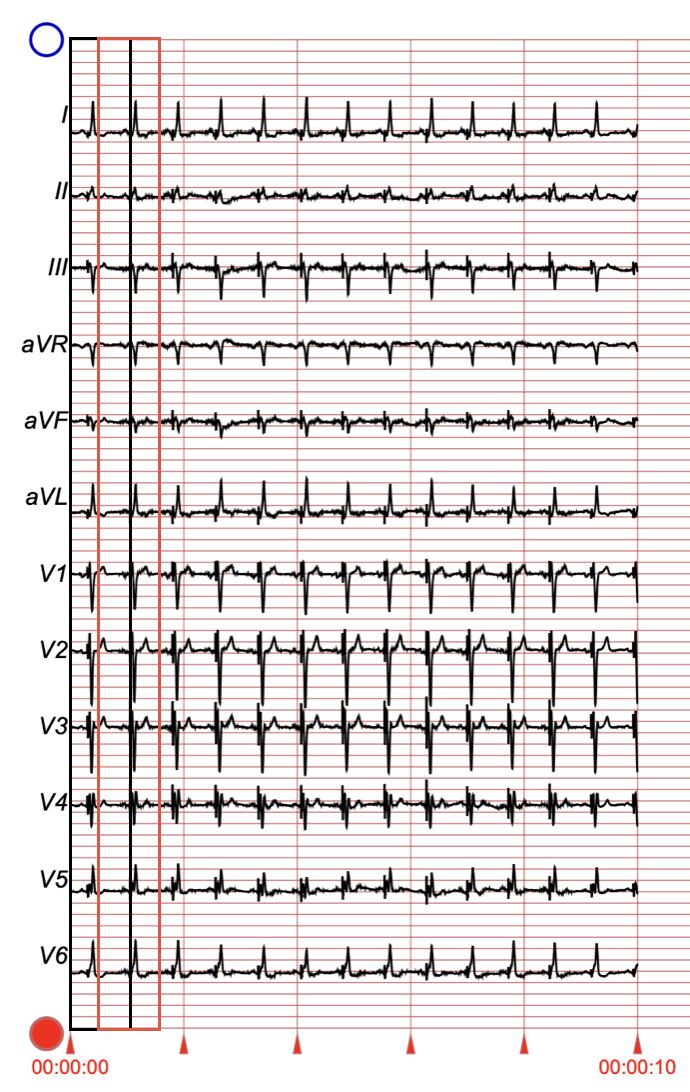}
  \caption{Example of a raw 12-lead ECG sample from the MIMIC-IV ECG dataset. The red and black boxes show consecutive windows extracted for training.}
  \label{fig:mimic_inputdata}
\end{figure}

Two ECG databases are used to train and validate each of the proposed models. Data curation was performed over all recordings labeled as Sinus Rhythm---normal heartbeats---from both datasets to ensure a refined and combined dataset containing only normal samples. The first dataset, PTB-XL \cite{ptbxl}, comprises twelve-lead recordings from PhysioNet \cite{physionet}, from which it was either cropped or padded to have an input data of ten-seconds segments to guarantee consistency within the second dataset. Each ECG sample is resampled at 500 Hz, providing 5,000 data points per lead, and a total of 8,900 normal samples were used with a 80/20 training and validation split (See Table \ref{tab:datasets_ptbxl_focus}). The second dataset, MIMIC-IV ECG \cite{mimic}, was introduced to perform real-world data augmentation after observing that the used architectures would perform better given a high-quality and high-quantity of ECG normal samples. After resampling all recordings to ten-second segments and filtering for valid normal rhythms, over 92,000 samples---labeled as Sinus Rhythm---were selected from more than 800,000 total samples available to include only those ECGs labeled uniquely as non-anomalous, excluding other data with additional diagnoses or ambiguous labeling systems. As shown in Figure \ref{fig:mimic_inputdata}, each sample consists of twelve-leads—each capturing ten seconds of the heart’s electrical activity.

\begin{figure*}[!ht]
  \centering
  \includegraphics[width=0.85\textwidth]{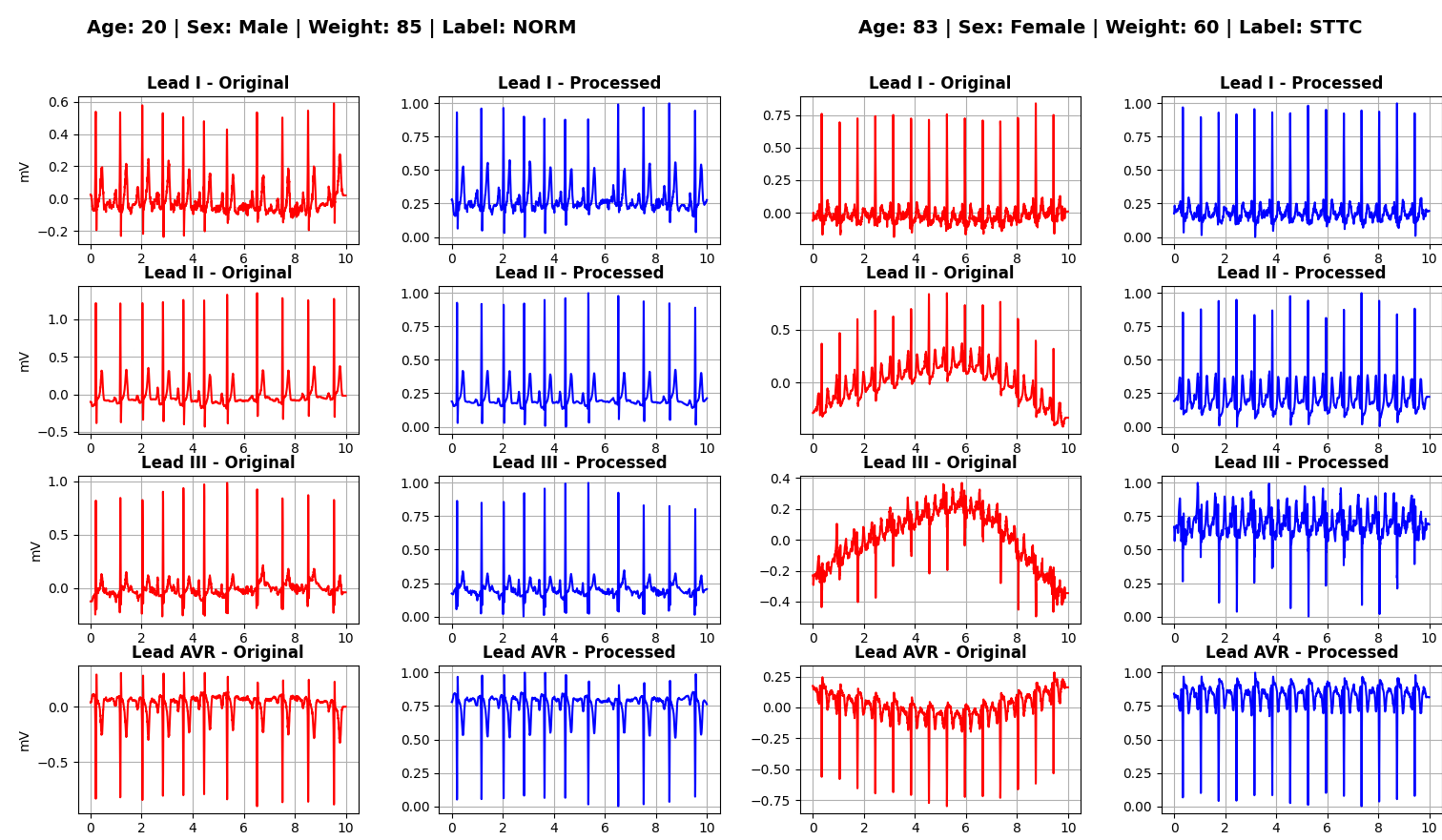}
  \caption{Comparison of non-processed and preprocessed ECG signals}
  \label{fig:plot_ecg_processed}
\end{figure*}

\begin{table}[ht]
\centering
\begin{tabular}{|l|r|r|}
\hline
\textbf{Characteristic} &
  \textbf{PTB-XL} &
  \textbf{MIMIC-IV ECG} \\ \hline
Total ECGs & \(21,799\) & \(>800,000\) \\ \hline
Sinus rhythm ECGs & \(\approx8,900\) & \(\approx92,000\) \\ \hline
Number of patients & \( 18,869\) & \(>160,000\) \\ \hline
Access & Public & Restricted \\ \hline
\end{tabular}
\caption{Input datasets overview}
\label{tab:datasets_ptbxl_focus}
\end{table}

\subsubsection{Window Segmentation}
\label{sec:window_segmentation}

Window segmentation is used during the experiments proposed in Section~\ref{sec:experiments}. This sample partition technique serves as a way to capture localized patterns within large ECG signals \cite{chauhan2015anomaly}. The training data set is matched by overlapping windows \(W\), where each window represents a fixed-length segment of the ECG signal: 

\begin{equation}
    W = \{ \mathbf{w}_1, \mathbf{w}_2, \ldots, \mathbf{w}_n \}, 
    \quad 
    \mathbf{w}_i \in \mathbb{R}^{L \times m}
    \label{eq:window_segments}
\end{equation}
where \(L\) is the number of ECG leads and each window 
\(\mathbf{w}_{i}\) is an \(L\times m\) matrix containing a fixed‐length segment from all leads.  The total number of windows per sample \(n\) is
\\
\begin{equation}
    n = \left\lfloor \frac{T - m}{s} \right\rfloor + 1,
    \quad
    s < m
    \label{eq:num_windows}
\end{equation}

In this case, \(T\) is the total length of the recording, \(m\) the window length (e.g., 500 samples), and \(s\) the hop size (e.g., 250 samples for 50 \% overlap). As shown in Figure~\ref{fig:mimic_inputdata}, the black and red-bordered rectangles over the first two seconds of an ECG recording illustrate how two windows are created and how the window segmentation is applied.

\subsubsection{Filtering and Normalization}
\label{sec:filtering}
After reviewing the literature and experimenting with different filtering and normalization configurations, the following techniques were chosen as optimal:

\begin{enumerate}
    \item ECG signals are cleaned using a combination of bandpass and notch filters \cite{preprocessing_filters, preprocessing_filters_2, preprocessing_filters_3}. Specifically, a 3rd-order Butterworth bandpass filter with cutoff frequencies of 0.5 Hz and 100 Hz is applied to remove baseline wander. Then, a notch filter centered at 60 Hz removes power-line interference. These filters are applied locally to each ECG lead to preserve signal quality and process them in a personalized way so that each lead is treated accordingly to its frequency distribution. 
    \item After filtering, the signals are normalized using z-score normalization \cite{zscore} in order to have a mean of 0 and a standard variation of 1 on a per-lead basis to ensure consistent amplitude scaling across the dataset, which facilitates effective training of machine learning models.
    \begin{equation}
    x_{t,i}^{(z)} = \frac{x_{t,i} - \mu_i}{\sigma_i + \varepsilon}
    \end{equation}
     %By normalizing the data using these values, each lead's signal is transformed to have an approximate mean of 0 and a standard variation of 1.
     where $i$ is the lead index, and the constant vector \(\epsilon\) is introduced to the denominator to prevent division by zero when the lead has zero variance.
\end{enumerate}

In Figure~\ref{fig:plot_ecg_processed}, the effects of filtering and normalization (in blue) are observable compared to the raw signals (in red). This preprocessing step reduces baseline drift and power-line interference. However, for anomalous samples, the same operations can also mask subtle deviations, consequently increasing the risk of false negatives.
For example, in the raw versus preprocessed comparison of the second sample, the low-amplitude ST-segment deviations visible in the raw signals are attenuated by the previously explained processing techniques. Consequently, there is a "trade-off between noise suppression and anomaly/morphology preservation" \cite{rodenas2018efficient}, as processing is an essential step before training a model to input non-noisy data, but it can also delete subtle anomalous patterns, thereby increasing the risk of false negatives.  

\subsection{Model Architectures}
\label{sec:model}
\subsubsection*{Convolutional Autoencoder (CAE)}

\begin{figure*}[!htb]
  \centering
  \includegraphics[width=0.8\textwidth]{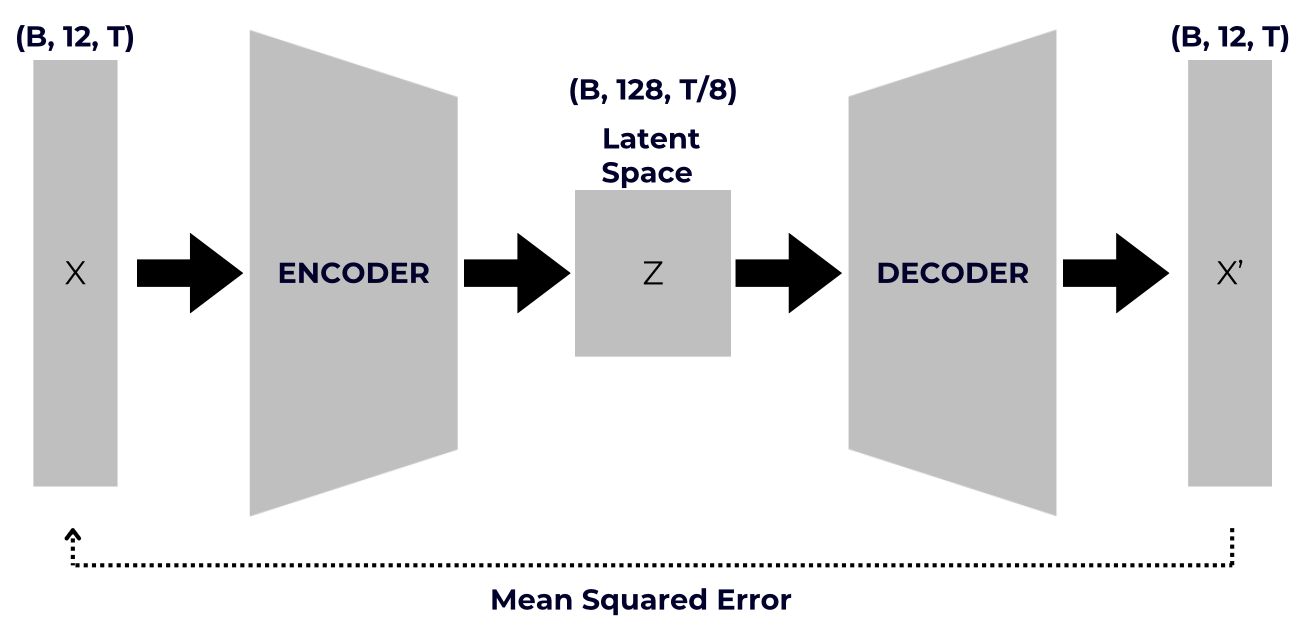}
  \caption{CAE architecture}
  \label{fig:CAE}
\end{figure*}

The proposed Convolutional Autoencoder, illustrated in Figure~\ref{fig:CAE}, is used as the baseline model in our comparative analysis and was adapted from open-source code \cite{rw_convae}. It processes twelve-lead ECG windows of shape $[B, 12, T]$, where $B$ is the batch size, 12 is the number of ECG leads per sample, and $T$ is the window length.

The model consists of two main blocks: an encoder that maps data $x$ into a latent representation $z$, and a decoder that maps this compressed space $z$ back into the original data representation $x'$ to perform a reconstruction of the original signal. The encoder compresses and reduces the temporal information using convolutional layers, while increasing channel depth. Afterwards, the decoder produces the inverse of this action, reconstructing $x'$ to match the original input shape. 

During training, the model minimizes the Mean-Squared Error (MSE) loss between each input sample and its reconstructed version. This loss function penalizes high-amplitude reconstruction deviations more heavily. In a 12-lead ECG, the largest per-sample amplitudes occur within the QRS complex, particularly at the R-peaks, where the signal changes sharply. Consequently, the autoencoder is driven to reproduce these segments with high fidelity, while small deviations in flatter regions (e.g., PR or ST segments) have a minor effect on the total loss.

\begin{figure*}[!htb]
  \centering
  \includegraphics[width=0.8\textwidth]{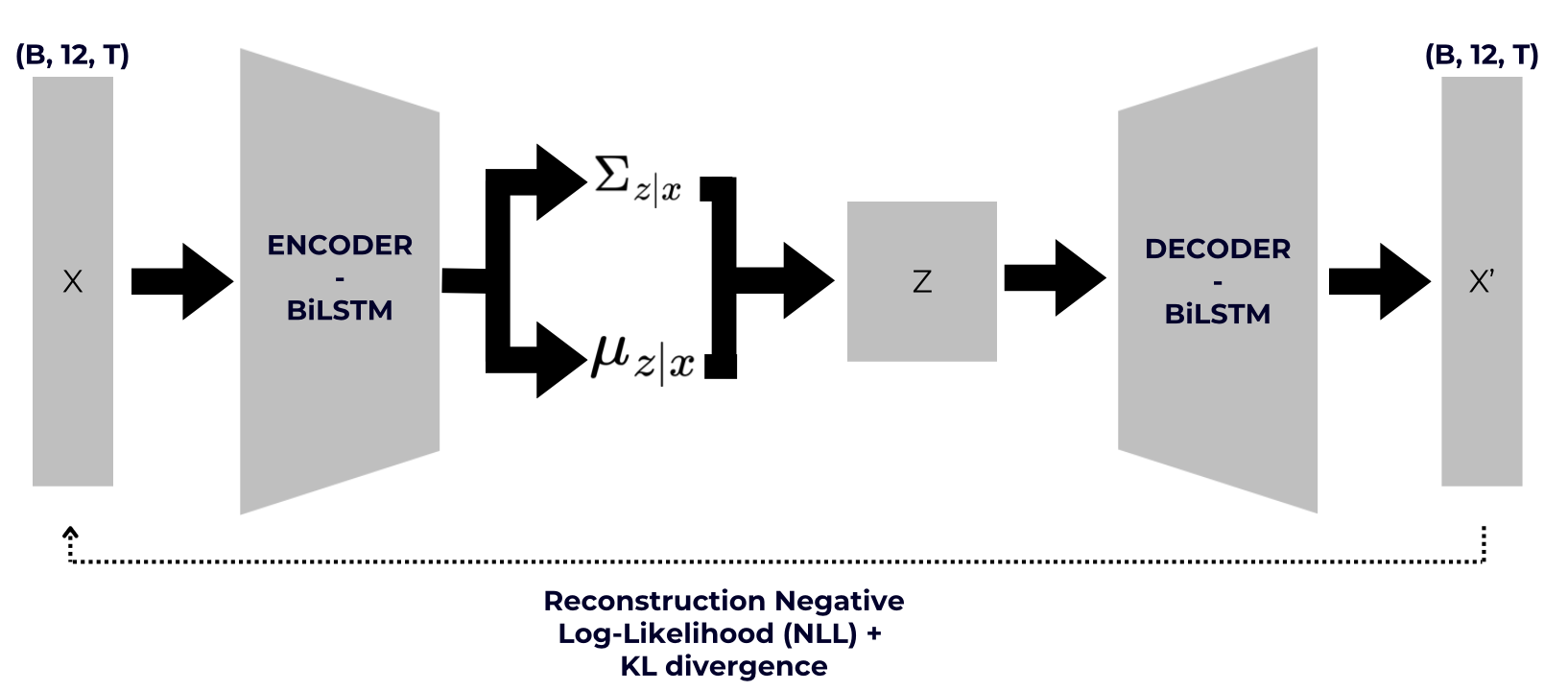}
  \caption{VAE-BiLSTM architecture}
  \label{fig:VAE}
\end{figure*}

\subsubsection*{\raggedright Variational Autoencoder Bidirectional Long Short-Term Memory (VAE-BiLSTM)}
The proposed VAE-BiLSTM model, illustrated in Figure~\ref{fig:VAE}, encodes each twelve-lead ECG window of shape $[B,12,T]$ (batch size $B$, leads $12$, timesteps $T$)---with a permutation to convert it to the required shape $[B,T,12]$--- with a bidirectional LSTM encoder block that generates a latent mean vector $\boldsymbol{\mu}_z\in\mathbb{R}^{B\times d}$ and a log-variance vector $\log\boldsymbol{\sigma}^2_z\in\mathbb{R}^{B\times d}$, where $d$ is the latent dimension. Afterwards, the reparameterization trick \footnote{The reparameterization trick is used in VAE models to enable gradient propagation with a differentiable transformation expressing a deterministic transformation (see Eq. \ref{eq:reparam_trick}).
%\begin{equation}
%\mathbf{z}
%  = \boldsymbol{\mu}_z
%    + %\exp\!\bigl(0.5\,\log\boldsymbol{\sigma}^2_z\bigr)\odot\boldsymbol{\epsilon},
%\qquad
%\boldsymbol{\epsilon}\sim\mathcal{N}(\mathbf{0},\mathbf{I})
%\label{eq:reparam_trick}
%\end{equation}
} is used to convert a latent sample $z$ from a non-differentiable sampling step to a differentiable one. Then,  $\mathbf{z}$ is then repeated $T$ times in a loop process and it is used as input to a unidirectional LSTM decoder, which outputs per-lead reconstruction means $\widehat{\mathbf{x}}_t$ and log-variances $\log\widehat{\boldsymbol{\sigma}}^2_t$, with tensors of shape $[B,12,T]$.

During training, the model minimizes the negative log-likelihood (NLL) of the output plus a Kullback-Leibler (KL) divergence multiplied by an annealed coefficient $\beta$. The architecture is inspired by the original VAE formulation of Kingma and Welling~\cite{kingma2013auto} and by the OmniAnomaly framework for time‐series anomaly detection~\cite{OmniAnomaly}.

This stochastic approach introduces variability in the latent representation, rather than relying on a point-wise reconstruction error---MSE---used in the previous deterministic model.

\subsubsection*{\raggedright VAE-BiLSTM with Multi-Head Attention (VAE-BiLSTM-MHA)}
\begin{figure*}[!htb]
  \centering
  \includegraphics[width=0.8\textwidth]{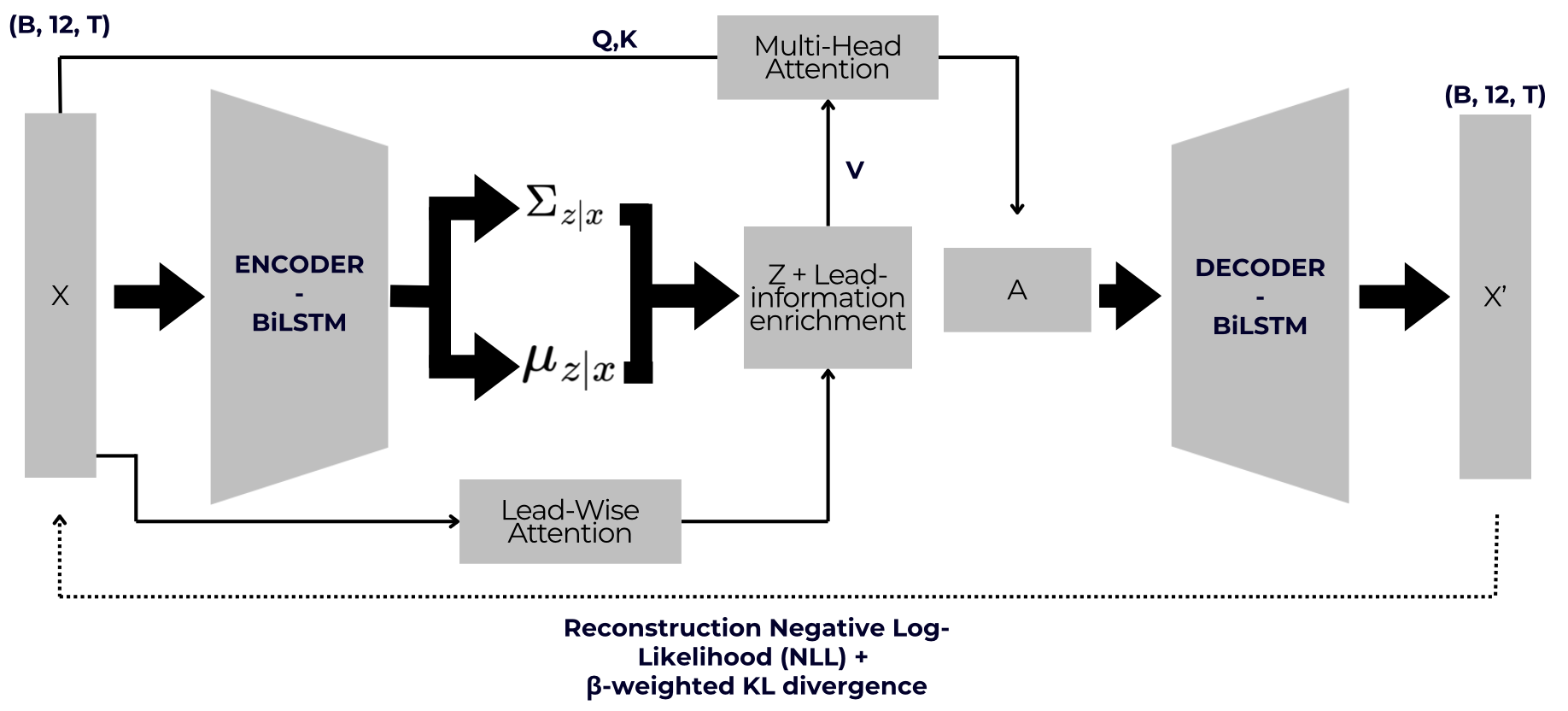}
  \caption{VAE-BiLSTM with Multi-Head Attention architecture}
  \label{fig:MAVAE}
\end{figure*}

Inspired by the original VAE architecture of Kingma and Welling \cite{kingma2013auto}, the OmniAnomaly time–series architecture \cite{OmniAnomaly}, and the MA-VAE design~\cite{correia2023mavae}, an extension of the previous model, VAE-BiLSTM, is presented in order to incorporate two attention mechanisms: a lead-wise attention to capture inter-lead dependencies and a multi-head attention to enhance the latent sequence representation as illustrated in Figure~\ref{fig:MAVAE}.

Each 12-lead window $\mathbf{x}\!\in\!\mathbb{R}^{B\times T\times 12}$ is encoded by two bidirectional BiLSTM layers, producing hidden states $\mathbf{h}\!\in\!\mathbb{R}^{B\times T\times 2h}$, where $h$ is the second LSTM’s hidden size. A fully connected layer (MLP) maps $\mathbf{h}$ to per-timestep latent mean and log-variance 
$\bigl[\boldsymbol{\mu}_z,\log\boldsymbol{\sigma}^2_z\bigr]\in\mathbb{R}^{B\times T\times 2d}$,  
with latent dimension $d$. Latent samples are then generated using the reparameterization trick \cite{kingma2013auto}:
\begin{equation}
\mathbf{z}
  = \boldsymbol{\mu}_z
    + \exp\!\bigl(0.5\,\log\boldsymbol{\sigma}^2_z\bigr)\odot\boldsymbol{\epsilon},
\qquad
\boldsymbol{\epsilon}\sim\mathcal{N}(\mathbf{0},\mathbf{I})
\label{eq:reparam_trick}
\end{equation}

To add inter-lead correlations, a lead-wise attention block embeds every $(B,T)$ slice of the input leads and applies a 4-head self-attention layer, resulting in lead context vectors $\tilde{\mathbf{h}}\!\in\!\mathbb{R}^{B\times T\times d}$. At the end, an enriched representation is obtained given the combination of the lead-wise attention with the latent space $\mathbf{z}^{\star}=\mathbf{z}+\tilde{\mathbf{h}}$.

Then, the latent $\mathbf{z}^{\star}$ is used as \emph{values} in a 8-head attention layer whose \emph{queries} and \emph{keys} are linear projections of the raw input window. This produces a context-aware sequence representation  
$\mathbf{A}\!\in\!\mathbb{R}^{B\times T\times d}$ that now encodes both global latent information and lead-specific saliency.

Finally, $\mathbf{A}$ is passed through two bidirectional LSTMs and an output MLP, providing per-sample reconstruction mean and logvar statistics. %$\bigl.
 
\subsection{Anomaly Detection Task}
\label{sec:anomaly_detection}
Given an unseen multi‐lead ECG sample of length $T$, the overlapping window segmentation is applied to obtain $n$ windows per sample (See Section~\ref{sec:window_segmentation} for the window process):

\begin{align}
W_x &= \bigl\{\mathbf{w}_1,\mathbf{w}_2,\dots,\mathbf{w}_n\bigr\}, \\[6pt]
n &= \left\lfloor \frac{T - m}{s} \right\rfloor + 1,
\quad s < m.
\end{align}
where
\begin{itemize}
  \item $T$ is the total length of the ECG recording in samples,
  \item $m$ is the length of each window in samples,
  \item $s$ is the stride between successive windows,
  \item $n$ is the total number of windows

\end{itemize}

Each window $\mathbf{w}_i$ is then passed through the trained anomaly detector model, which produces a per‐window anomaly score

\begin{equation}
s_i = \mathcal{S}\left( \mathbf{w}_i \right)
\label{eq:score}
\end{equation}
where the scoring function $\mathcal{S}$ depends on the model architecture:
\begin{itemize}
    \item For the CAE model, the anomaly score for each window is computed as the mean‐squared reconstruction error:
\begin{equation}
s_i^{\mathrm{CAE}} = \|\mathbf{w}_i - \hat{\mathbf{w}}_i\|_2^2
\label{eq:anomaly_score}
\end{equation}

where the MSE measures point-wise differences between the input and its reconstruction ~\cite{zhang2021unsupervised}. Consequently, windows with large reconstruction errors are considered as anomalous signals.  

    \item For the VAE‐BiLSTM model, each window’s anomaly score is computed by the negative evidence lower bound (ELBO \cite{kingma2013auto}), where the MSE is used as the reconstruction term, which can be derived from maximizing the Gaussian log-likelihood:
\begin{equation}
\begin{aligned}
s_i^{\mathrm{VAE\text{-}BiLSTM}} 
&= \|\mathbf{w}_i - \hat{\mathbf{w}}_i\|_2^2 \\
&\quad+ \mathrm{KL}\bigl(q_\phi(z \mid \mathbf{w}_i) \,\|\, p(z)\bigr)
\end{aligned}
\label{eq:vae_bilstm}
\end{equation}

where the first section measures the MSE reconstruction error and the second section is the KL divergence between the approximate posterior \(q_\phi(z\mid \mathbf{w}_i)\) and the prior \(p(z)\), enforcing a regularized latent space. Consequently, a window's high score can lead to an unlikely latent representation and a poor reconstructed sample.  
    \item For the VAE-BiLSTM-MHA model, we define the anomaly score as an attention‐weighted ELBO:

\begin{equation}
\begin{aligned}
s_i^{\mathrm{VAE\text{-}BiLSTM \text{-}MHA}}
&= \underbrace{\sum_{t=1}^{m}\alpha_{i,t}\,\bigl\|\mathbf{w}_{i,t} - \hat{\mathbf{w}}_{i,t}\bigr\|_2^2}_{\text{Attention‐weighted reconstruction}} \\[6pt]
&\quad+\;\underbrace{\mathrm{KL}\bigl(q_\phi(z\mid \mathbf{w}_i)\,\|\,p(z)\bigr)}_{\text{Latent regularization}}
\end{aligned}
\label{eq:vae_mha_score}
\end{equation}

where \(\alpha_{i,1},\dots,\alpha_{i,m}\) considers normalized attention weights produced per each window, the \(\|\mathbf{w}_{i,t} - \hat{\mathbf{w}}_{i,t}\|_2^2\) determines the MSE reconstruction error and the other section regularizes the latent space against the prior as in the case of VAE-BiLSTM above. %denotes the KL divergence between the approximate posterior \(q_\phi(z\mid \mathbf{w}_i)\) and the prior \(p(z)\), enforcing a regularized latent space.

The attention-weighted reconstruction term is based on the assumption that the self-attention block inside the model ``attends'' to those time steps where there are relevant variations. Consequently, low-information segments contribute less than in salient zones regarding the MSE. The correlation between high-attention regions and high reconstruction error will be illustrated in the visualizations presented in Section \ref{sec:results}.
\end{itemize}

\paragraph{Decision Rule}
Once each window $\mathbf{w}_i$ has been assigned a score $s_i$, a single score is obtained for the entire recording by taking the mean: 
\begin{equation}
S = \frac{1}{n} \sum_{i=1}^{n} s_i
\label{eq:global_score}
\end{equation}
As explained in Section~\ref{sec:problem}, a threshold $\tau$ is established during validation. Then, the final anomaly decision is given by
\[
\hat{y} =
\begin{cases}
\text{anomalous}, & S > \tau,\\
\text{normal},    & S \le \tau.
\end{cases}
\]
If the average score $S$ exceeds $\tau$, the signal is labeled as anomalous; otherwise, it is labeled as normal. This simple rule leverages the validation‐tuned threshold to balance sensitivity and specificity.

\section{Experiments}
\label{sec:experiments}

The three proposed autoencoder-based models (CAE, VAE-BiLSTM, and VAE-BiLSTM-MHA) are evaluated against four literature baselines listed below, which also report results on the same dataset. The comparison is made under the closest possible conditions:
\begin{itemize}
  \item ConvAE \cite{rw_convae}  
  \item VAE–AE Hybrid \cite{rw_vae_ae}  
  \item AE + Peak Detection \cite{rw_ae_peak_detection} 
  \item MSGformer \cite{ji2024msgformer}
\end{itemize}

Below, Tables \ref{tab:hp_cae}–\ref{tab:beta_vae_mha} summarize the hyperparameter configurations for each of our proposed models. All parameters were selected manually, aided by visual analysis.

\begin{table}[H]
\centering
\caption{CAE Hyperparameters}
\begin{tabular}{ll}
\toprule
\textbf{Parameter}       & \textbf{Value}                 \\
\midrule
Window size             & $500$                            \\
Stride                  & $0.5\times$ window size ($250$)  \\
Learning rate           & $1\times10^{-3}$               \\
Number of epochs        & $100$                            \\
Criterion               & MSE loss                       \\
Optimizer               & Adam                           \\
\bottomrule
\end{tabular}
\label{tab:hp_cae}
\end{table}

\begin{table}[H]
\centering
\caption{VAE–BiLSTM Hyperparameters (excluding $\beta$)}
\begin{tabular}{ll}
\toprule
\textbf{Parameter}       & \textbf{Value}                        \\
\midrule
Window size             & $500$                                   \\
Stride                  & $0.5\times$ window size ($250$)        \\
Learning rate           & $5\times10^{-3}$                      \\
Number of epochs        & $100$                                   \\
Latent dimension        & $64$                                    \\
Hidden dimension        & $128$                                   \\
Criterion               & Recon.\,$+$\,KL                       \\
Optimizer               & Adam                                  \\
\bottomrule
\end{tabular}
\label{tab:hp_vae_bilstm}
\end{table}

\begin{table}[H]
\centering
\caption{VAE–BiLSTM-MHA Hyperparameters (excluding $\beta$)}
\begin{tabular}{ll}
\toprule
\textbf{Parameter}       & \textbf{Value}                             \\
\midrule
Window size             & $500$                                      \\
Stride                  & $0.5\times$ window size ($250$)            \\
Learning rate           & $1\times10^{-4}$                           \\
Number of epochs        & $100$                                      \\
Latent dimension        & $64$                                       \\
Hidden dimension        & $128$                                      \\
Number of attention heads & $8$                                      \\
Dropout (encoder)       & $0.1$                                      \\
Gaussian noise (input)  & $\sigma = 0.01$                            \\
Criterion               & Attn.\,$+$\,Recon.\,$+$\,KL                \\
Optimizer               & Adam                                       \\
\bottomrule
\end{tabular}
\label{tab:hp_vae_mha}
\end{table}

\begin{table}[H]
\centering
\caption{$\beta$‐Annealing Schedule for VAE–BiLSTM}
\begin{tabular}{ll}
\toprule
\textbf{Epoch } ($t$)    & \textbf{$\beta_t$}                       \\
\midrule
$t \le 10$               & $\displaystyle \beta_t = \frac{t}{10}\,$ \\
$t > 10$                 & $\displaystyle \beta_t = 1.0$            \\
\bottomrule
\end{tabular}
\label{tab:beta_vae_bilstm}
\end{table}

\begin{table}[H]
\centering
\caption{$\beta$‐Scheduling for VAE-BiLSTM–MHA (cyclical ramp from $10^{-8}$ to $10^{-2}$ over epochs [11–100])}
\begin{tabular}{ll}
\toprule
\textbf{Epoch } ($t$)    & \textbf{$\beta_t$}                                                                           \\
\midrule
$t \le 10$               & $\beta_t = 10^{-8}$                                                                           \\
$10 < t \le 100$         & $\displaystyle \beta_t = 10^{-8} + \frac{t - 10}{90}\,[10^{-8} , 10^{-2}]$                     \\
\bottomrule
\end{tabular}
\label{tab:beta_vae_mha}
\end{table}

\subsection{Datasets}
\label{sec:datasets}

The primary evaluation dataset is the publicly available CPSC 2018 challenge corpus \cite{cpsc}, which comprises 12-lead ECG recordings acquired at 500~Hz for a nominal duration of 10~s per patient. The dataset is balanced in terms of gender (approximately 50~\% female, 50~\% male).  For the binary anomaly detection task, each recording has been relabeled as either \emph{Sinus Rhythm} (i.e., normal) or \emph{Anomalous}, obtaining over 2,000 ECG samples---each recording from a unique patient---resulting to over 2,000 patients \cite{li2021automatic}.

The preprocessing pipeline remained identical to that used for the MIMIC-IV ECG data, except for the notch filter, whose center frequency was lowered from 60~Hz to 50~Hz to match the mains frequency of the region in which the new dataset was recorded.

\begin{table*}[!ht]
\centering
\caption{Performance Comparison of the Proposed and (cited) Baseline Models on the CPSC Dataset}
\label{tab:benchmark_models}
%\setlength{\tabcolsep}{12pt}      % reduce horizontal padding (default ~6pt)
%\renewcommand{\arraystretch}{0.9} % reduce vertical padding (default 1.0)
%\scriptsize                      % reduce font size
\begin{tabular}{lccccc}
\toprule
\textbf{Model} & \textbf{Prec.} & \textbf{Rec.} & \textbf{F1} & \textbf{AUROC} & \textbf{AUPRC} \\
\midrule
CAE                       & 0.64 & 0.82 & 0.72 & 0.77 & 0.80 \\
VAE-BiLSTM                & 0.70 & 0.76 & 0.73 & 0.78 & 0.81 \\
\textbf{VAE-BiLSTM-MHA}   & \textbf{0.75} & \textbf{0.85} & \textbf{0.80} & \textbf{0.80} & \textbf{0.81} \\
\midrule
ConvAE \cite{rw_convae}           & –    & –    & 0.79 & 0.80 & –    \\
VAE–AE Hybrid \cite{rw_vae_ae}    & –    & –    & –    & 0.65 & –    \\
AE + Peak Detection \cite{rw_ae_peak_detection} & – & 0.56 & 0.56 & 0.74 & – \\
MSGformer \cite{ji2024msgformer}  & –    & –    & \textbf{0.847} & \textbf{0.88} & – \\
\bottomrule
\end{tabular}
\end{table*}

To ensure consistency with the training preprocessing and windowing (Section~\ref{sec:filtering}), the testing set passed through identical preprocessing steps---bandpass filtering to remove baseline wander and power‐line interference, followed by z-score normalization per lead--- except for the notch filter, in which case its maximum frequency was lowered from 60~Hz to 50~Hz to match the main frequency of the region in which the CSPC dataset was recorded. Additionally, a selection of 10~s duration segments at 500~Hz was performed. This curation and preprocessing pipeline ensures that both training and test sets follow a standard preprocessing approach, as used in several research articles \cite{preprocessing_filters, preprocessing_filters_2}.

\subsection{Evaluation Metrics}
\label{sec:evaluation_metrics}

To evaluate the performance of the proposed anomaly detection model, a set of standard metrics commonly used in anomaly detection literature was selected, including precision, recall, F1 score, AUPRC, and AUROC.

\paragraph{Precision, Recall, and F1 Score:}  
The primary evaluation metrics are precision (\( P \)), recall (\( R \)), and F1 score (\( F_1 \)):

\begingroup\small
\begin{equation}
  P = \frac{TP}{TP + FP}, \quad
  R = \frac{TP}{TP + FN}, \quad
  F_1 = 2\,\frac{P\,R}{P + R}
\end{equation}
\endgroup

where  \( TP \), \( FP \), and \( FN \) are the number of True Positives (correctly detected anomalies), False Positives (normal samples incorrectly labeled as anomalies), and False Negatives (missed anomalies), respectively.

\paragraph{AUPRC:}  
The Area Under the Precision-Recall Curve determines the model's ability to correctly detect anomalies (\(TP\)) while avoiding false alarms (\(FP\)). 

\paragraph{AUROC:}  

The Area Under the Receiver Operating Characteristic Curve evaluates the trade-off between the true positive rate (\(TPR = \tfrac{TP}{TP+FN}\)) and the false positive rate (\(FPR = \tfrac{FP}{FP+TN}\)) across different thresholds. That is to say, a higher AUROC value shows a stronger discriminative ability between healthy and anomalous ECG samples. 

\subsection{Results}
\label{sec:results}

Table~\ref{tab:benchmark_models} summarizes the anomaly detection performance of all models evaluated on the CPSC 2018 test set. The results of the proposed models are obtained under a unified evaluation process, whereas the baseline scores are directly extracted from the respective original publications and reflect their reported evaluation conditions. Below we detail the findings of the experiments with each model.

\subsubsection*{Convolutional Autoencoder (CAE)}
The convolutional autoencoder achieves a solid performance on the quantitative anomaly detection task. Additionally, the interpretability of this model is obtained through the saliency map\footnote{A saliency map is a visualization that shows which parts of an input the model considers most important.} using the model's MSE reconstruction error.

\begin{figure*}[!ht]
  \centering
  \includegraphics[width=0.9\textwidth]{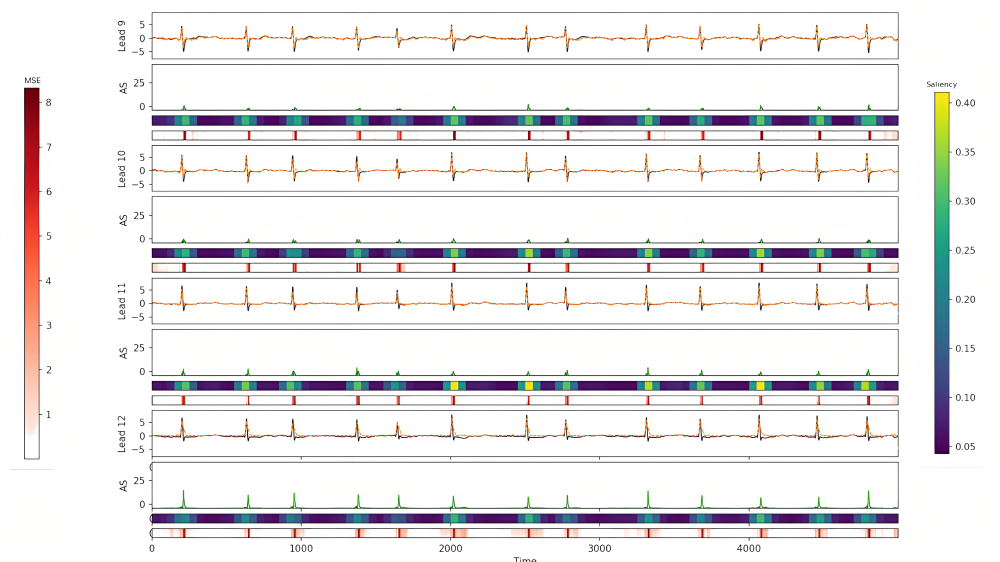}
  \caption{CAE interpretability of a normal ECG sample}
  \label{fig:CAE_VISUALIZATION}
\end{figure*}

Figure \ref{fig:CAE_VISUALIZATION} illustrates an original normal ECG signal (black) and its reconstruction (orange) per each lead, with three extra information: the per-window anomaly score (``A''), the saliency heat-map, and the MSE reconstruction error values. As observed in the figure, the highest MSE values and the darkest saliency are located in every QRS. It means that the model has learned to focus on these rapid and large changes in the signals. More importantly, even though these spikes obtain the highest MSE error, the global anomaly score remains below the defined threshold, meaning that the CAE model interprets the spikes within the normal morphology.

\subsubsection*{Variational Autoencoder BiLSTM (VAE-BiLSTM)}

Compared with the CAE model, the VAE–BiLSTM achieves higher F1, AUROC, and AUPRC, since its stochastic latent variables force the model to learn representations that generalize from normal-only training data. However, in early experiments the model suffered from \emph{posterior collapse}, where the approximate posterior collapsed onto the prior as formulated in Equation~\ref{eq:posterior_collapse}:

\begin{equation}
q_\phi(\mathbf{z}\mid\mathbf{x}) \;\approx\; p(\mathbf{z})
\label{eq:posterior_collapse}
\end{equation}

At collapse, the KL divergence term fell to nearly zero, the encoder outputs $\boldsymbol{\mu}_z\approx\mathbf{0}$ and $\log\boldsymbol{\sigma}^2_z\approx\mathbf{0}$, and the decoder didn't receive information about the latent space $z$, converting it into a deterministic auto‐encoder (since $\beta\,\mathrm{KL}\approx0$). 

To solve this problem, a technique called cyclical KL annealing \cite{correia2023mavae} was applied (see Table~\ref{tab:beta_vae_mha}), increasing $\beta$ over time starting at epoch 11---letting the model also learn a well-reconstructed normal manifold---over a 10-epoch schedule and then repeating the cycle. This technique allowed the model to distribute importance between reconstruction and regularization phases---MSE and KL divergence accordingly. A dropout was also added for further regularization.  

Overall, this combined strategy prevented posterior collapse as each latent dimension maintained a minimum of information flow, and the posterior parameters \(\mu_z\) and \(\log\sigma^2_z\) stayed separated from the prior values, thus, the encoder did not collapse to \(\mu_z=0\) and \(\log\sigma^2_z=0\).

\begin{figure*}[!t]
  \centering
  \includegraphics[width=0.7\textwidth]{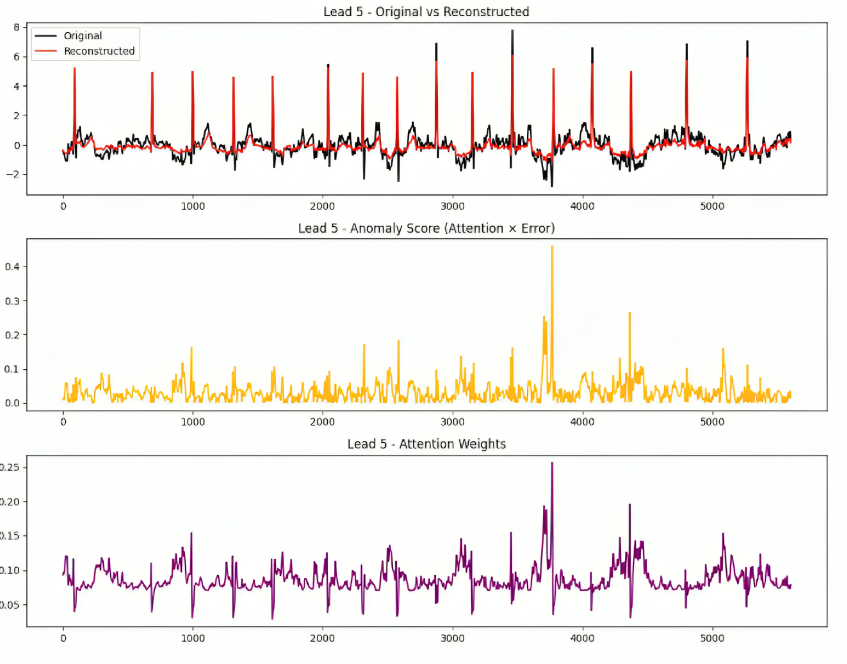}
  \caption{VAE-BiLSTM-MHA interpretability of an anomalous ECG sample}
  \label{fig:VAE_EXPLAINABILITY1}
\end{figure*}

\begin{figure*}[!t]
  \centering
  \includegraphics[width=0.7\textwidth]{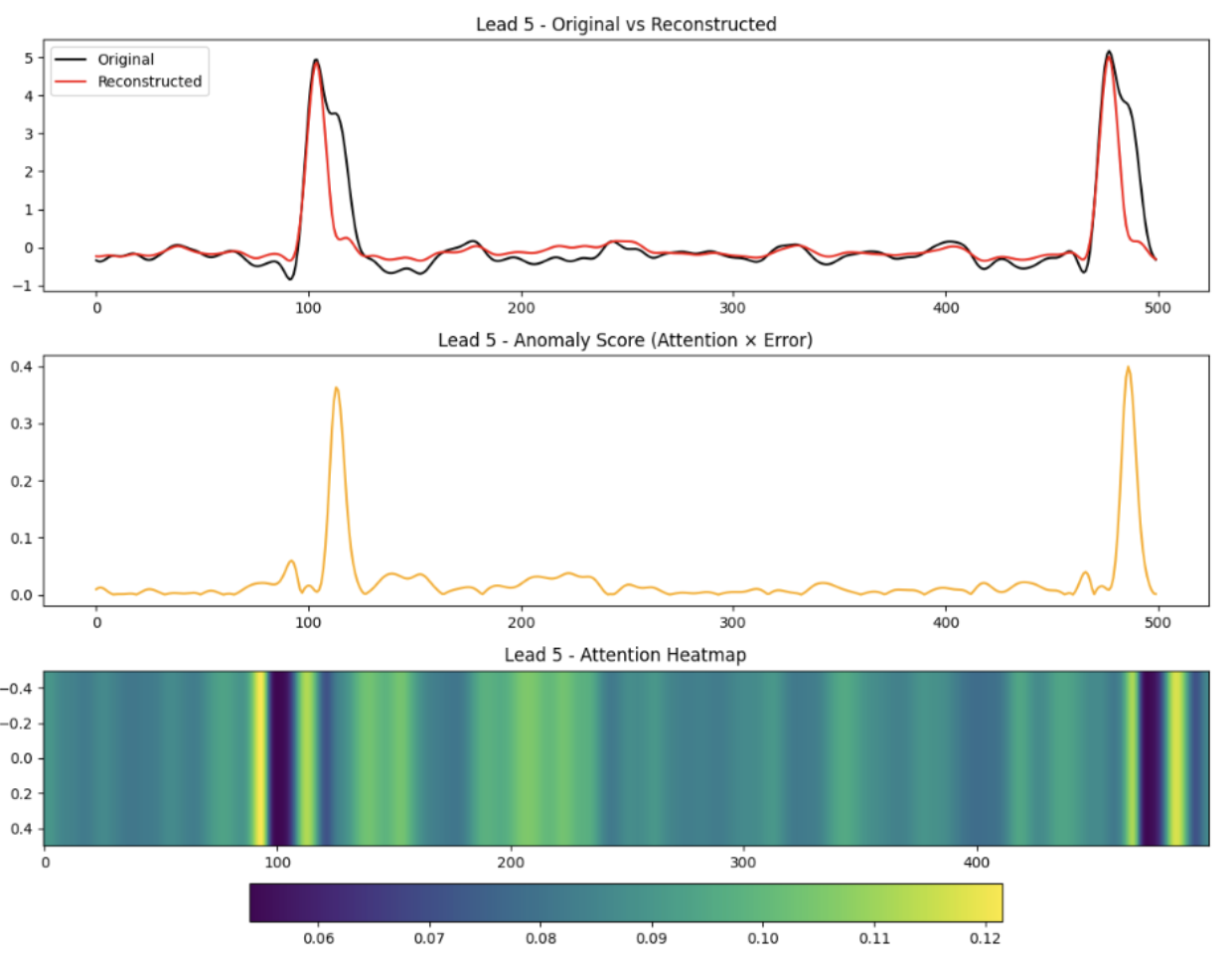}
  \caption{VAE-BiLSTM-MHA interpretability of Figure~\ref{fig:VAE_EXPLAINABILITY1}, focused on a window size of 500 samples}
  \label{fig:VAE_EXPLAINABILITY2}
\end{figure*}

\subsubsection*{\raggedright VAE-BiLSTM with Multi-Head Attention (VAE-BiLSTM-MHA)}

The last model achieves improved performance on the quantitative task, showing higher Precision, Recall, F1-score, and AUROC compared to the previous models. Additionally, its interpretability is enhanced through the attention mechanisms. 
Figure~\ref{fig:VAE_EXPLAINABILITY1} shows a 12‐lead anomalous ECG sample (black original vs. red reconstruction), together with the per‐window anomaly score in yellow and the attention weights in purple. It illustrates that between time steps 3,500 and 4,000 the model produces higher reconstruction error, higher anomaly scores, and increased attention weights in the same section. This correlation indicates that the model not only struggles to reconstruct the signal in that specific region, but also assigns more importance to that region through its attention mechanism. The combination of reconstruction error and attention probably means that the model is highlighting a segment it both finds difficult to reproduce and finds it relevant for the overall result. 

Figure~\ref{fig:VAE_EXPLAINABILITY2} focuses on a 500 timestep window of the previous anomalous ECG sample, showing the original vs reconstructed signal, the anomaly score and the attention weights within a heatmap. The peaks in reconstruction error and anomaly score are strongly correlated and they are located in QRS peaks. Interestingly, the heatmap shows that the model does not focus exactly on the sharp error peaks, but rather on the segments before and after them. This indicates that the model considers the surrounding waveform morphology, rather than focusing only on the sharp peaks. Such behavior leads to a more informative explanation, as subtle variations before and after the QRS complex can indicate anomalous patterns. 

\subsection{Dashboard}
\label{sec:dashboard}

A local web application is implemented as the last objective of this work, it uses the Dash framework within Plotly integration to create visualizations of these last interpretable models---both CAE and VAE-BiLSTM-MHA. As it is observed in Figure \ref{fig:dashboard}, the user interface offers a simple and personalized framework where you can upload an ECG signal in standard formats---\texttt{.npy} or \texttt{.dat}/\texttt{.hea}. Afterwards, clicking on ``Analyze ECG Sample'', the interface will render the 12-lead ECG signal, predict an anomaly decision, and provide as well as a detailed plot to interpret model's reasoning. 

The dashboard also offers configuration possibilities such as choosing the leads that you wanna visualize, choosing between a CAE and a VAE-BiLSTM-MHA models, and deciding the anomaly threshold---though it is not recommend to alter the default threshold value as it is computed on the validation set (Section~\ref{sec:problem}). In the main plot section, there are several displays per each lead. The original signal (black) and the reconstructed signal (red) are found in the first graphic of each lead. Then, there is another graphic that shows the anomaly score (orange) in each timestep. Moreover, there are two horizontal heat-map bars that compute the lead-wise attention weights and the point-wise MSE per leads along time steps. 

\section{Conclusions}
\label{sec:conclusions}

In this work, a comparative analysis was conducted among three autoencoder-based architectures for anomaly detection in multivariate 12-lead electrocardiogram (ECG) time-series data. Specifically, a convolutional autoencoder (CAE), a variational autoencoder with bidirectional long short-term memory (VAE-BiLSTM), and a VAE-BiLSTM augmented with multi-head attention (VAE-BiLSTM-MHA) architectures were evaluated on the CPSC 2018 dataset. The results demonstrated that the proposed novel approach VAE-BiLSTM-MHA outperformed the other models, achieving an AUPRC of 0.81 and a recall of 0.85. Beyond quantitative performance, this model also exhibited qualitative strengths regarding the interpretability of anomaly diagnosis. An additional contribution of this work was the development of a user-friendly interactive web application that provides a simple way to observe ECG signals and indicate possible anomaly deviations with interpretable graphics, helping users identify and understand cardiac anomalies. 

{
  \captionsetup[figure*]{aboveskip=0pt, belowskip=4pt}%
  \begin{figure*}[!t]
    \centering
    \includegraphics[width=1\linewidth]{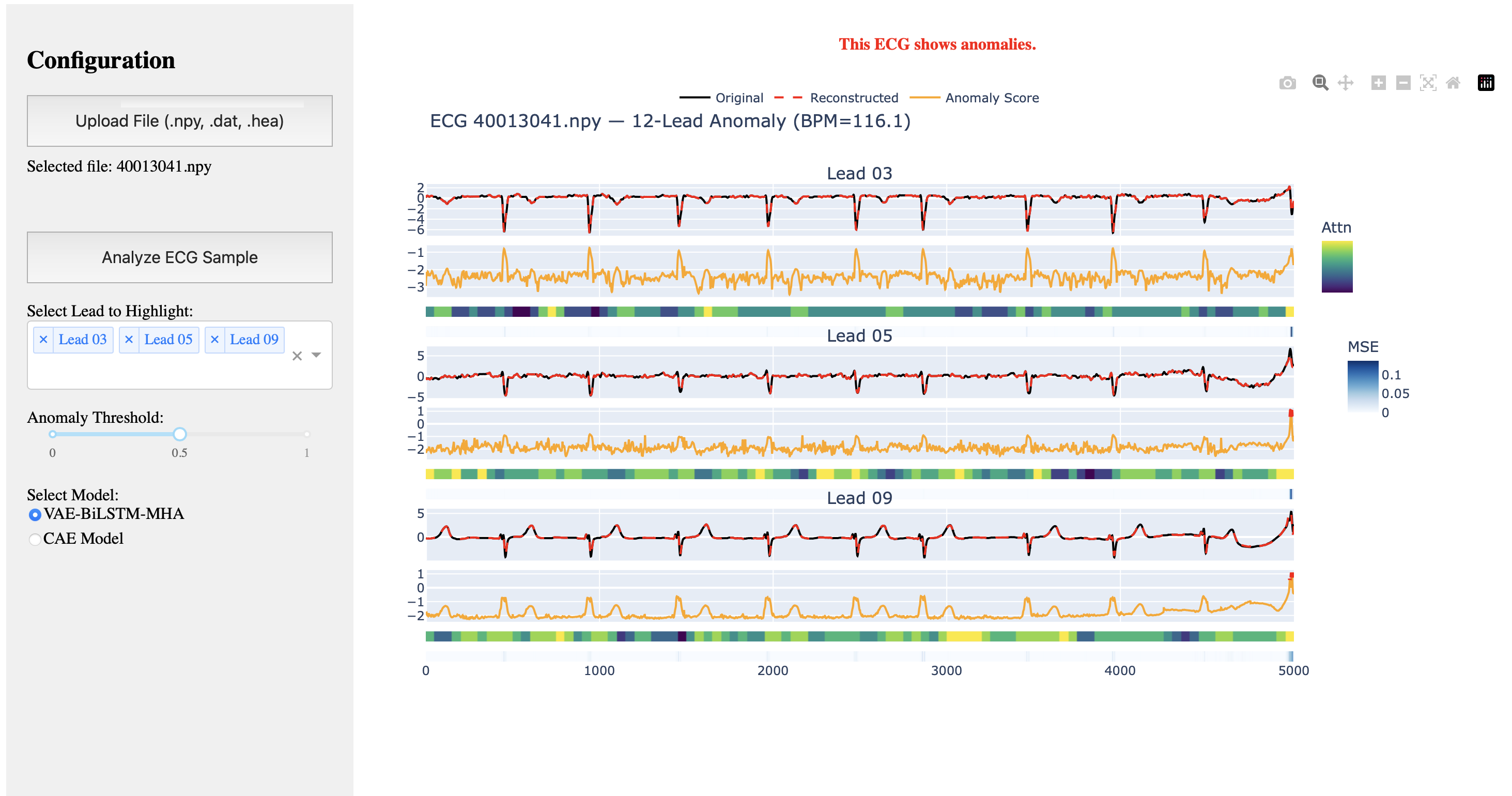}
    \caption{Interactive Dashboard}
    \label{fig:dashboard}
  \end{figure*}
}
%\FloatBarrier

Despite these advances, limitations are present in this study. First, the evaluation on a single benchmark dataset with customized configurations defined within this work may limit the generalizability of the results. Second, the models were trained and evaluated in an offline setting, without facing real-time ECG monitoring challenges. Finally, the absence of medical professionals in the process of model creation and its evaluation restricted the ability to fully assess the clinical interpretability of the models' outputs. 

Future research directions include extending the current VAE-BiLSTM-MHA framework for real-time, online anomaly detection in collaboration with medical professionals, to facilitate deeper clinical insights into ECG data and its anomalies. Such partnerships would foster innovation by combining technical expertise with the domain knowledge of healthcare practitioners. Additionally, exploring advanced architectures such as Transformer-based models and integrating additional patient-related data with ECG signals may improve both the accuracy of anomaly detection and the interpretability of diagnostic outcomes. 

\section*{Software Availability}
\urlstyle{same}
The source code has been made publicly available at: \\ \url{https://github.com/marcgarreta/12-lead-ECG-AD}

\section*{Acknowledgments}
The authors would like to extend their special thanks to \mbox{Universitat Autònoma de Barcelona} for providing the computational resources that supported this research.

\bibliographystyle{IEEEtran}
\bibliography{IEEEabrv,main}

\end{document}